\documentclass{article}
\usepackage{ijcai22}

\usepackage{times}
\usepackage{soul}
\usepackage{url}
\usepackage[hidelinks]{hyperref}
\usepackage[utf8]{inputenc}
\usepackage[small]{caption}
\usepackage{graphicx}
\usepackage{amsmath}
\usepackage{amsthm}
\usepackage{booktabs}
\usepackage{algorithm}
\usepackage{algorithmic}
\usepackage{booktabs}       % professional-quality tables
\usepackage{multirow}

\title{Selective Output Smoothing Regularization: Regularize Neural Networks by Softening Output Distributions}

\author{
Xuan Cheng$^1$
\and
Tianshu Xie$^1$\and
Xiaomin Wang\And
Minghui Liu\and
Jiali Deng\and
Ming Liu$^2$\\
\affiliations
$^1$Authors contributed equally to this work.\\
$^2$Author to whom correspondence should be addressed.\\
}

\begin{document}

\maketitle

\begin{abstract}
  In this paper, we propose Selective Output Smoothing Regularization, a novel regularization method for training the Convolutional Neural Networks (CNNs). Inspired by the diverse effects on training from different samples, Selective Output Smoothing Regularization improves the performance by encouraging the model to produce equal logits on incorrect classes when dealing with samples that the model classifies correctly and over-confidently. This plug-and-play regularization method can be conveniently incorporated into almost any CNN-based project without extra hassle. Extensive experiments have shown that Selective Output Smoothing Regularization consistently achieves significant improvement in image classification benchmarks, such as CIFAR-100, Tiny ImageNet, ImageNet, and CUB-200-2011. Particularly, our method obtains 77.30$\%$ accuracy on ImageNet with ResNet-50, which gains 1.1$\%$ than baseline (76.2$\%$). We also empirically demonstrate the ability of our method to make further improvements when combining with other widely used regularization techniques. On Pascal detection, using the SOSR-trained ImageNet classifier as the pretrained model leads to better detection performances.
\end{abstract}

\section{Introduction}
Improving the performance of deep convolutional neural networks (CNNs) has long been a question of great interest in a wide range of fields. Recent improvements in network structure and computer hardware have indeed brought promising performances, but they have also entailed increased risks of overfitting. To combat with this issue, a considerable literature has grown up around the themes of data augmentation and regularization. Taken together, these techniques have focused on various details of training, ranging from the inputs (Data Augmentations~\cite{yun2019cutmix,devries2017improved,cubuk2018autoaugment,lim2019fast,zhong2020random,xie20cut-thumbnail}), to labels(Label Smoothing~\cite{szegedy2016rethinking} and OLS~\cite{zhang2021delving}), to parameters (Weight Decay), to depths (Stochastic Depths~\cite{huang2016deep}), to neurons (Dropout), to feature maps (Batch Normalization, DropBlock~\cite{ghiasi2018dropblock}) and so on. However, far little attention has been paid to regularizing the output distribution of the network.

Deep networks are known to exhibit a tendency to make high confidence predictions. Most studies addressing this issue have only focused on the uncertainty issue on out-of-distribution images~\cite{guo2017calibration,hendrycks2018deep,moon2020confidence} while ignoring the possible resultant negative effects on original training. Diving deeper into this, we observe that in the early stage of training, a large number of ``easy samples'' have already been correctly classified with high confidence. Due to the non-linearity of the cross-entropy loss, this observation suggests that loss values and gradients of these samples will be overwhelmed by samples with a relatively small probability, resulting in a dispensable role of high-probability samples.

To tackle this problem, we propose a simple yet effective idea which help add supplemental instructions for these samples. Since the model has already classified these samples correctly, we choose to take measures on the incorrect labels. According to the maximum entropy principle~\cite{jaynes1957information}, with no further knowledge of incorrect classes provided, there is no need to favor any of incorrect classes over the others. To this end, we enforce the model to produce equal logits on incorrect classes for samples that gets a high probability. Specifically, we add a mean squared error loss that represents the difference between the original output and the generated desired output to the cross-entropy loss. Under the joint supervision of these two losses, the model obtains more highly discriminative features and thereby enlarges the inter-class feature differences. As our experiments will show, we observe the consistent and significant accuracy boosts using various architectures on multiple datasets. It implies that many deep CNNs have been suffering from this problem and have not been unleashed with their full powers yet, where our method can help. We refer to this method as Selective Output Smoothing Regularization, or SOSR.

SOSR shares certain similarity with Label Smoothing~\cite{szegedy2016rethinking} and Confidence Penalty~\cite{pereyra2017regularizing}. While both adopting the average operation, Label Smoothing modifies the \textit{label distribution} through reducing the expected label by $\epsilon$ and distributing it on average. Our method instead modifies the \textit{output distribution} by evening the unexpected logits regardless of the expected logits. Besides, the motivation for Label Smoothing is to prevent the model from being too confident while our method focuses on providing supplemental instructions when the model has already been too confident for some samples. With respect to Confidence Penalty, despite both being affected by the principle of maximum entropy, Confidence Penalty operates on \textit{all} the outputs to prevent peaked distributions while our method instead differentiates the samples and only exert pushes on \textit{partial} samples that the model is too confident about. More importantly, extensive experiments will show that our method not only consistently outperforms Label Smoothing and Confidence Penalty but also makes further improvements when combined with these two methods.

To summarize, the proposed method has the following main advantages:
\begin{itemize}
	\item It’s a simple but effective method that can be easily implemented with few lines of code. Without any interference in learning strategy and network architecture, it acts more like a “plug-and-play component” that we can easily “assemble” into the project.
	\item Albeit simple, various architectures with our method consistently show nice performance boosts on various recognition benchmarks.
	\item It’s complementary to most widely used data augmentation and regularization methods.
\end{itemize}

%-------------------------------------------------------------------------
\section{Our Method}
\subsection{A closer look at high confidence predictions}
\label{observe}
%Our method's main motivation comes from an intuitive idea that different samples shall influence the model to different degrees. For starters, we describe the cross-entropy loss quantitatively. Suppose we use softmax as an activation function after the last layer of the network, the probability $p_k$ that the model assigns to the $k$-th class should be $p_k=\frac{e^{o_k}}{\sum_{l=1}^{K}e^{o_l}}$, where $o_k$ represents the un-normalized logits the model assigns to the $k$-th class, $K$ is the total number of the predicted classes. For a network trained with cross-entropy loss, we are interested in minimizing the expected value between the true targets $q_k$ and the probability $p_k$. Mathematically, it can be expressed as:
%\begin{equation}
	%H(\textbf{\textit{q}},\textbf{\textit{p}}) = \sum_{k=1}^{K}-q_k\log(p_k)
%\end{equation}
%Further consider the case where there is only one ground-truth label $y$, so that $q_y=1$ and $q_k=0$ for all $k\neq y$. In this case, the value of cross-entropy loss becomes $H=-\log(p_y)$. One notable property of this loss is that it is a %nonlinear function with respect to $p_y$.  
% To this end, we expect to find a general method to detect the over-confident samples and encourage the model to produce equal logits on the incorrect classes when dealing with over-confident samples. 

We first describe some empirical observations on the model providing high confidence predictions. These observations serve as a key intuition for designing and understanding our algorithm. 

An active line of research has shown that neural networks tend to make high confidence predictions, frequently produced by softmaxes because softmax probabilities are computed with the fast-growing exponential function~\cite{hendrycks2016baseline}. Such phenomenon has raised great concerns particularly on how can classifiers obtain awareness of uncertainty when fed with new kinds of input~\cite{guo2017calibration,hendrycks2018deep,moon2020confidence}, but we seldom think about how this would affect the ongoing training process. Still due to the exponential function, a slight variation in softmax probabilities results in a huge difference in the value of loss. For example, the value of cross-entropy loss in 0.3 probability is hundreds of times as much as the value in 0.99 probability. That being said, when summed over all the samples, samples with a relatively high probability actually make minimal difference to the model due to values of loss and gradients being overwhelmed by samples with a relatively small probability.

\begin{figure}[h]
	\begin{center}
		\includegraphics[width=0.9\columnwidth, height=1\linewidth]{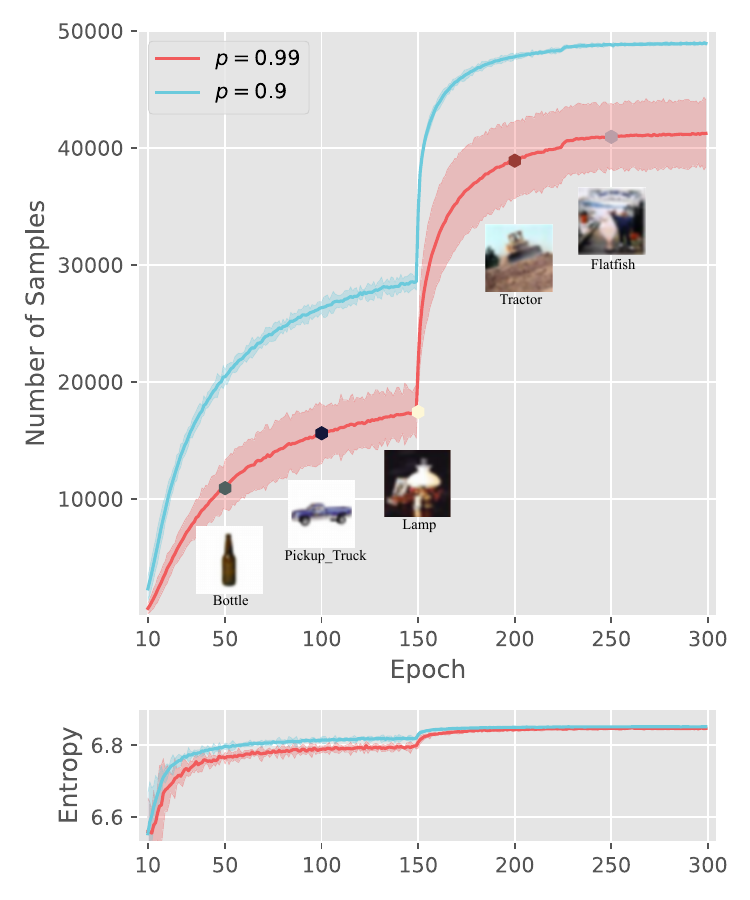}
	\end{center}
	\caption{\textbf{Top:}The number of training images that get a probability greater than $p$ on the correct label. The experiments were conducted on CIFAR-100 using ResNet-110. \textbf{Bottom:}Average information entropy of these samples(that are correctly classified with high confidence) in black-and-white. Shaded areas represent three times of standard deviation on 4 trials.}
	\label{baseline}
\end{figure}

The top plot of Figure~\ref{baseline} shows the number of samples that get high probabilities on the correct classes during training, using ResNet-110 on CIFAR-100. In the bottom plot, we depict the average information entropy of these samples in black-and-white. For an image, the entropy characterizes the complexity of possible pixel values, thereby being ideal for the measure of the image’s difficulty. We observe that:(1) networks tend to first learn the ``easy samples'' and defer learning ``hard samples'' (samples with high entropy). The improvement in model capability allows more and more hard samples to be classified correctly and highly confidently by the network, which lines up with recent findings on training dynamics of deep neural networks~\cite{arpit2017closer,lee2019robust,nam2020learning}. Another interesting observation is that (2) a large number of easy samples have already been recognized by the network with high confidence in the early stage of training. Take $p=0.99$ for example, when training for 300 epochs, over 10000 images reach a probability greater than 99$\%$ on the 50-th epoch, suggesting at least 1/5 of training images barely provide useful information in the next 250 epochs. 

%-------------------------------------------------------------------------
\subsection{Design}
\noindent\textbf{Motivation. } Based on the findings in Section~\ref{observe}, increased network capacity would result in decreasing number of samples being effectively trained, even in the early stage of training. Thus, measures should be adopted for samples, in which the model assigns a high value of probability to the correct class. For brevity, we refer to this kind of samples here and after as “over-confident samples”. Now that the model has already been able to correctly classify the over-confident samples, trying to deal with the incorrect classes seems to be a good choice. Intuitively speaking, a perfect model should be able to identify a sample with the correct class accurately and, more importantly, without confusion with any other classes. In other words, logits the model assigns to the incorrect classes should be equal. It can also be interpreted from the principle of maximum entropy~\cite{jaynes1957information}. Since the current state of knowledge only provides information about the correct class, there is no reason to favor any incorrect class over the other incorrect classes. By forcing the model to generate equal output logits on incorrect classes for over-confident samples, we not only better train the model by artificially producing more information about over-confident samples, but the model also learns to focus on more distinguishing and prominent features in one class, keeping it free from the disruption of other classes.

\noindent\textbf{Formulation. }Here we describe the Selective Output Smoothing Regularization algorithm in detail. Consider a typical image classification setting where we have a training dataset consisting of tuples of images and labels $(x,y)\in X\times Y$ and we are interested in optimizing a network to generalize to unseen data. The goal of SOSR is to force the network to assign the same logits to incorrect classes for over-confident samples. To do that, we first generate a copy of the output $O$ as the desired output $\tilde{O}$. Then we convert the output into normalized probabilities with softmax function. For each training example $x$ in the output $O$, if its maximum normalized probability is greater than a given threshold $P$ and corresponding to the ground-truth label, it will be considered over-confident. For each over-confident sample, we modify its copy in the desired output $\tilde{O}$ by changing its unexpected logits to their mean values. Finally, we add the mean squared error loss between the original output $O$ and the desired output $\tilde{O}$ to the cross-entropy loss $L_{CE}$ with weight $\beta$, so that
\begin{equation}
	\mathcal{L} = L_{CE} + \beta \frac{1}{MK}\sum_{m=1}^{M}\sum_{k=1}^{K}(\tilde{O} - O)^2
\end{equation}
where $M$ represents the batchsize and $K$ is the total number of the predicted classes. 

\begin{figure}[h]
	\begin{center}
		\includegraphics[width=0.8\columnwidth, height=0.5\linewidth]{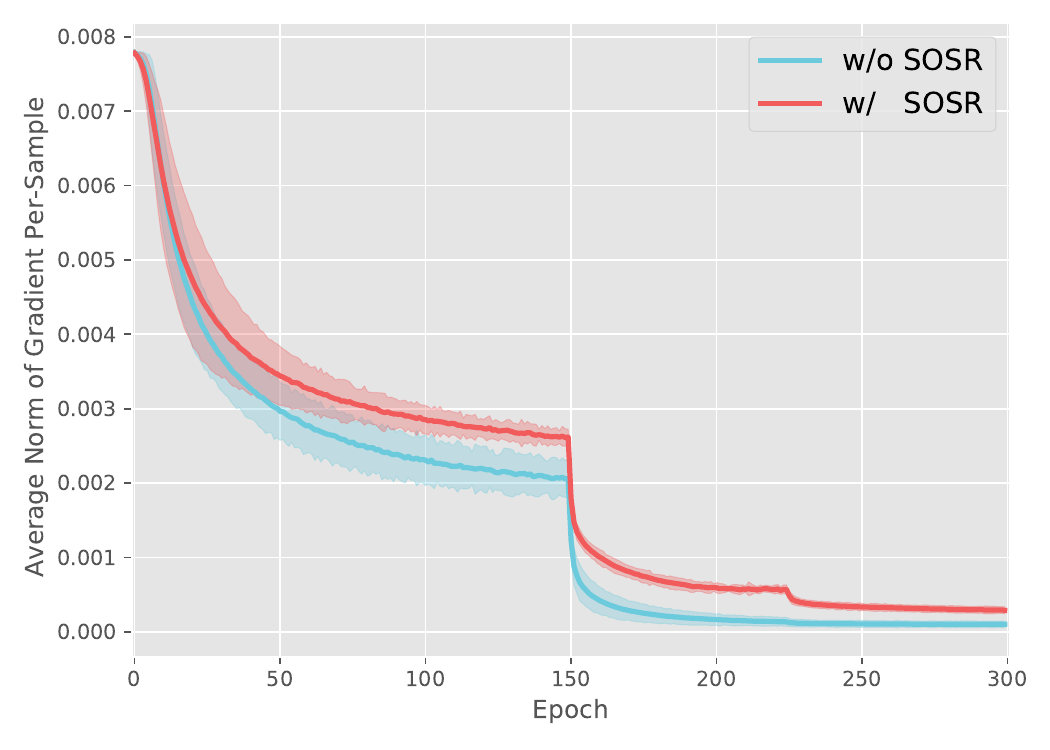}
	\end{center}
	\caption{Average norm of the gradient for the outputs per sample of SOSR against its counterpart as training proceeds on CIFAR-100. Shaded areas represent three times of standard deviation on 4 trials. SOSR results in a slower downward curve.}
	\label{grad}
\end{figure}

\noindent\textbf{Analysis. }Notably, for samples that are not over-confident, we in fact remain their copies in $\tilde{O}$ unchanged. Conversely, for each copy of over-confident samples, we remain its maximum logit unchanged but revise other logits to the average values of all the unexpected logits of it. Then, under the supervision of the mean squared error loss, we are able to enforce the model to assign equal logits on the unexpected label for over-confident samples. By doing this, we provide further instructions for mere over-confident samples when their original gradients have been overwhelmed by other non-over-confident samples. As we can see from Figure~\ref{grad}, we visually compare the average norm of the gradient of each sample for SOSR against the vanilla setting as training progressed. We can observe that the average norm of gradient of baseline drops quickly since increasing number of samples fail to provide enough gradients, whereas the gradients of SOSR exhibit a slower downward trend due to the further instructions imposed on over-confident samples.

There are only two parameters in the algorithm, which are probability threshold $P$ and weight $\beta$. The threshold $P$ marks the boundary between normal samples and the over-confident samples. It determines precisely how many training samples need to be processed in every iteration. Specifically, with the decreased value of the threshold $P$ comes with the increased number of images to be processed. Note that the performance may degrade when you reduce the value of $P$ too much. It is not surprising to understand this effect in terms of erasure of information. When applying our method, the model is encouraged to treat each incorrect class as equally probable. Hence an excessive reduction in value of $P$ may impair the model’s generalization ability to some extent. The weight $\beta$ controls the strength of SOSR. The detailed empirical analysis of the effect of both parameters will be shown in ablation studies present in Supplementary Materials. 

Code-level details and illustration of the algorithm pipeline are provided in Supplementary Materials. It can be seen that SOSR is easy to implement with few lines and can be incorporated into any network architecture. It’s also the purpose we design it first. Acting like a “\textit{plug-and-play component}” that can be easily assembled into any classification-based project, our method enjoys the property that it can bring consistent and steady improvements with or without other techniques involved.
%-------------------------------------------------------------------------
\section{Experiments}
In this section, we apply Selective Output Smoothing Regularization to multiple tasks to evaluate its performance and capability. Note that we mainly report the highest Top-1 classification accuracy over the full training process for a fair comparison, to avoid the lower accuracy due to overfitting by the end of training. All the experiments are implemented and evaluated on 8$\times$1080Ti GPU with \textit{Pytorch}. If not specified, all results reported are averaged over 4 runs.

\subsection{Classification Accuracy on Different Benchmarks}
We empirically demonstrate the effectiveness of our method on a series of benchmark datasets: CIFAR-100~\cite{krizhevsky2009learning}, Tiny ImageNet, ImageNet~\cite{russakovsky2015imagenet} and Caltech-USCD Birds-200-2011~\cite{WahCUB_200_2011}.

\noindent\textbf{CIFAR-100:} CIFAR-100 consists of 60000 $32\times32$ color images, representing 100 classes of natural scene objects. We incorporate SOSR on 8 architectures: Vgg\cite{simonyan2014very}, ResNet\cite{he2016deep}, RegNet~\cite{Radosavovic2020}, DenseNet\cite{huang2017densely}, Wide Residual Networks\cite{zagoruyko2016wide}, PyramidNet\cite{han2017deep}, MobileNet\cite{sandler2018mobilenetv2}, SENet\cite{hu2018squeeze}. Each baseline and its SOSR counterpart are trained with standard data augmentation strategies such as random flipping and random cropping, following the common practice~\cite{huang2016deep}. Full implementation details are present in Supplementary Materials.

\begin{table}[h]
	\begin{center}
		\begin{small}
			\begin{tabular}{p{4cm}|p{1.46cm}p{1.46cm}}
				\toprule[1.3pt]
				\midrule
				Model & Baseline & + Ours\\
				\midrule
				Vgg19$\_$bn		  &71.55{\scriptsize $\pm$0.11}   & \textbf{72.18}{\scriptsize $\pm$0.05}\\
				\midrule
				ResNet-56         &73.61{\scriptsize $\pm$0.06}   & \textbf{74.70}{\scriptsize $\pm$0.13} \\
				ResNet-110        &74.53{\scriptsize $\pm$0.09}   & \textbf{75.6}{\scriptsize $\pm$0.10}\\
				\midrule
				RegNetX\_200MF     &76.54{\scriptsize $\pm$0.05}   & \textbf{77.16}{\scriptsize $\pm$0.03} \\
				\midrule
				DenseNet-100-12   &77.22{\scriptsize $\pm$0.05} &\textbf{77.92}{\scriptsize $\pm$0.03}\\
				\midrule
				WRN-28-10         &78.84{\scriptsize $\pm$0.02}  & \textbf{79.95}{\scriptsize $\pm$0.07}\\
				\midrule
				PyramidNet-110-48   &76.88{\scriptsize $\pm$0.04}  & \textbf{77.23}{\scriptsize $\pm$0.08}\\
				PyramidNet-110-84   &79.08{\scriptsize $\pm$0.02}  & \textbf{79.41}{\scriptsize $\pm$0.07}\\
				\midrule
				MobileNet           &68.15{\scriptsize $\pm$0.03}  & \textbf{70.22}{\scriptsize $\pm$0.15}\\
				MobileNetV2         &72.48{\scriptsize $\pm$0.05}  & \textbf{73.19}{\scriptsize $\pm$0.09}\\
				\midrule
				SE-ResNet-110              &76.12{\scriptsize $\pm$0.32}  & \textbf{77.14}{\scriptsize $\pm$0.14}\\
				\bottomrule[1.3pt]
			\end{tabular}
		\end{small}
	\end{center}	
	\caption{Accuracy ($\%$) with and without SOSR using different architectures on CIFAR-100.}
	\label{cifar-results}
\end{table}

We display the results of validation accuracy in Table~\ref{cifar-results}. We fix $P=0.99$ for all the experiments. Results indicate that models trained with our method consistently achieve significant improvement, demonstrating that it is applicable to various CNN architectures. For ResNet-56 and ResNet-110, our method improves the accuracy by 1.09$\%$ and 0.89$\%$, respectively. Notably, \textit{these results are achieved using the same threshold $P$, thus not representing the best performance our method can get}. We draw attention to the fact that our method yields these performance improvements even when applied to excessively complex models that consist of dozens of times as many parameters as those small networks.

\noindent\textbf{Tiny ImageNet:} We then evaluate our approach on a medium-scale dataset. Tiny ImageNet dataset is a subset of the ImageNet dataset with 200 classes. It has images with 100k and 10k samples for training and validation sets, respectively. All images are 64×64 colored ones. Implementation details are provided in Supplementary Materials.We still fix $P=0.99$ for both experiments. Results with and without our method are given in Table~\ref{tiny-results}. We observe that our method outperforms the baseline accuracy by 0.88$\%$ with ResNet-110 model.
\begin{table}[h]
	\begin{center}
		\begin{small}
			\begin{tabular}{p{3.5cm}p{1.5cm}p{1.5cm}}
				\toprule[1.3pt]
				\midrule
				Model & Baseline($\%$) & + Ours($\%$)\\
				\midrule
				ResNet-56         &60.22{\scriptsize $\pm$0.16}   & \textbf{60.85}{\scriptsize $\pm$0.04} \\
				ResNet-110        &62.59{\scriptsize $\pm$0.10}   & \textbf{63.47}{\scriptsize $\pm$0.25} \\
				\midrule
				\bottomrule[1.3pt]
			\end{tabular}
		\end{small}
	\end{center}	
	\caption{Accuracy with and without SOSR using different architectures on Tiny ImageNet.}
	\label{tiny-results}
\end{table}

\noindent\textbf{ImageNet:} Finally, we evaluate our method on a large-scale dataset, the ILSVRC 2012 classification dataset. It contains 1.2 million training images and 50000 validation images labeled with 1000 categories. Due to the lack of time and computational resources, each experiments here were only conducted twice. We report the best performance of our method and baselines with ResNet-18 and ResNet-50 model during training in Table~\ref{imagenet-results}. The experimental results show that our method has a top-1 accuracy of 77.30$\%$, which is 1.1$\%$ higher than the baseline training. Both the improvements on ResNet-18 and ResNet-50 demonstrate the effectiveness of our method on large-scale datasets. We also note that, the superior performance on ResNet-50 implies that our method leads to better performance improvement than Label Smoothing, which only achieves a consistent improvement of about 0.2$\%$, as quoted in~\cite{szegedy2016rethinking}.

\begin{table}[h]
	\begin{center}
		\begin{small}
			\begin{tabular}{p{3.5cm}p{1.5cm}p{1.5cm}}
				\toprule[1.3pt]
				\midrule
				Model & Top-1($\%$) & Top-5($\%$)\\
				\midrule
				ResNet-18         &70.45{\scriptsize $\pm$0.04}           & 90.01{\scriptsize $\pm$0.03} \\
				ResNet-18+SOSR    &\textbf{71.58}{\scriptsize $\pm$0.07}  & \textbf{90.33}{\scriptsize $\pm$0.05} \\
				\midrule
				ResNet-50         &76.20{\scriptsize $\pm$0.01}           & 92.97{\scriptsize $\pm$0.03}\\
				ResNet-50+SOSR    &\textbf{77.30}{\scriptsize $\pm$0.02}           & \textbf{93.43}{\scriptsize $\pm$0.01}\\
				\midrule
				\bottomrule[1.3pt]
			\end{tabular}
		\end{small}
	\end{center}	
	\caption{Validation accuracy with and without our method using two networks on ImageNet}
	\label{imagenet-results}
\end{table}

\noindent\textbf{CUB-200-2011:} Despite the remarkable progress CNN has made for general image classification, fine-grained image classification still remains to be challenging. We further test the effectiveness of our method on the most widely used fine-grained benchmark: Caltech-USCD Birds-200-2011. CUB-200-2011 contains 11788 images with 200 categories of birds. We use the default training/test split for experiments, which gives us around 30 training examples per class. We train the model for 95 epochs with ResNet-50 as our backbone architecture. The learning rate is set to 0.001, which decays by a factor of 10 every 30 epochs. We fixed $P=0.99$ when training.

Table~\ref{cub-result} shows the performance of SOSR. We first construct our baseline model over ResNet-50 network pre-trained by a regular process on ImageNet. By implementing SOSR, we construct a new model using a ResNet-50 network pre-trained with our method as the backbone architecture. Surprisingly, the SOSR pre-trained model successfully leads to a better performance by achieving an 85.92$\%$ top-1 accuracy. Besides, the addition of SOSR to this new model once again brings improvement in the new baseline, achieving 86.18$\%$ accuracy. The superior performance of SOSR and SOSR pre-trained model combination proves the ability of our method to obtain more highly discriminative features even for fine-grained image classification.

\begin{table}[h]
	\begin{center}
		\begin{small}
			\begin{tabular}{p{3.5cm}p{1.5cm}p{1.5cm}}
				\toprule[1.3pt]
				\midrule
				Backbone Network & Baseline($\%$) & + Ours($\%$)\\
				\midrule
				ResNet-50         &85.37{\scriptsize $\pm$0.03}   & \textbf{85.94}{\scriptsize $\pm$0.01} \\
				ResNet-50(SOSR)    &85.92{\scriptsize $\pm$0.05}   & \textbf{86.18}{\scriptsize $\pm$0.12} \\
				\midrule
				\bottomrule[1.3pt]
			\end{tabular}
		\end{small}
	\end{center}	
	\caption{Accuracy with and without our method on CUB-200-2011. ResNet-50 denotes the ResNet-50 network that is pre-trained with the regular process on ImageNet. ResNet-50 (SOSR) denotes ResNet-50 pre-trained with the addition of our method.}
	\label{cub-result}
\end{table}

\subsection{Performance over Other Regularization Methods}
To explore whether Selective Output Smoothing Regularization is complementary to other data augmentation and regularization methods, we have evaluated the combination of our method and some widely used techniques. We briefly describe the settings for baseline augmentation schemes. We selected the Cutout size of $8\times8$ based on the results of~\cite{devries2017improved}. We set the label smoothing parameter value and the confidence penalty weight to 0.1 to work best. The implementations of AutoAugment and Fast AutoAugment were based on the publicly available code. For CutMix, since it had two labels used for one generated sample, we changed the criterion for judging over-confident samples from the comparison of the maximum probability and the given threshold to comparison between the sum probability of two labels and the threshold. When modifying the desired output, we remained both the logits of two labels unchanged and revised other logits to their mean values.

\noindent\textbf{Combinations on CIFAR-100:} We first performed our experiments on CIFAR-100. We set $P=0.99, \beta=0.5$ for all the experiments. The quantitative results are given in Table~\ref{combinationoncifar}. Regardless of the architectural modification, the combination of our method and any other regularization methods tend to outperform using those alone. Note that, CutMix has significantly outperformed the baseline. Thus it’s remarkable for our method to gain such improvements when combining with Cutmix, considering our method itself is simple.
\begin{table}[h]
	\begin{center}
		\begin{small}
			\begin{tabular}{lllc}
				\toprule[1.3pt]
				\midrule
				Model                      & Method            & Acc. & + Ours \\ 
				\midrule
				\multirow{7}{*}{ResNet-56}  & +Cutout            & 74.64   & \textbf{75.17} \\    
				& +CutMix            & 76.58   & \textbf{76.88} \\
				\cmidrule(r){2-4}
				& +AutoAugment           & 76.28   & \textbf{76.77} \\
				& +Fast AutoAugment      & 75.96   & \textbf{76.37} \\
				\cmidrule(r){2-4}
				& +SOSR              & 74.70   & --\\
				& +LabelSmoothing       & 73.94   & \textbf{74.32} \\ 
				& +ConfidencePenalty & 73.81   & \textbf{74.22} \\
				\midrule
				\multirow{7}{*}{ResNet-110} & +Cutout            & 75.61   & \textbf{75.91} \\  
				& +CutMix            & 78.83   & \textbf{79.52} \\
				\cmidrule(r){2-4}
				& +AutoAugment           & 78.08   & \textbf{78.67} \\
				& +Fast AutoAugment      & 77.63   & \textbf{78.05} \\
				\cmidrule(r){2-4}
				& +SOSR              & 75.42   & --\\
				& +LabelSmoothing       & 74.94   & \textbf{75.44} \\ 
				& +ConfidencePenalty    & 74.75   & \textbf{75.35} \\
				\midrule
				\bottomrule[1.3pt]
			\end{tabular}
		\end{small}
	\end{center}
	\caption{Accuracy ($\%$) over the combination of our method and other techniques on CIFAR-100. We list the performance of SOSR itself for easier comparison with Label Smoothing and Confidence Penalty.}
	\label{combinationoncifar}
\end{table}

\noindent\textbf{Combinations on Tiny ImageNet:} We also performed experiments on Tiny ImageNet. We still fix $P=0.99, \beta=0.5$. For the sake of briefness, we only combined our methods with CutMix. As seen in Table~\ref{combinationontiny}, same phenomenons that the addition to the CutMix method brought better performance are observed. Due to the lack of computational resources, we could not combine our method with other techniques on ImageNet. Nevertheless, given that the combination of our method and other regularization techniques worked well on both CIFAR-100 and Tiny ImageNet, we can assume that the combination can successfully lead to better performance on ImageNet.
\begin{table}[h]
	\begin{center}
		\begin{small}
			\begin{tabular}{p{3.5cm}p{1.5cm}p{1.5cm}}
				\toprule[1.3pt]
				\midrule
				Model & Baseline($\%$) & + Ours($\%$)\\
				\midrule
				ResNet-56 + CutMix        &61.22{\scriptsize $\pm$0.03}   & \textbf{62.36}{\scriptsize $\pm$0.01} \\
				ResNet-110 + CutMix       &66.13{\scriptsize $\pm$0.01}   & \textbf{66.58}{\scriptsize $\pm$0.02}\\
				\midrule
				\bottomrule[1.3pt]
			\end{tabular}
		\end{small}
	\end{center}
	
	\caption{Accuracy ($\%$) over the combination of our method and other techniques on Tiny ImageNet.}
	\label{combinationontiny}
\end{table}

\noindent\textbf{Combinations on CUB-200-2011:} Table~\ref{combinationoncub} shows the results of SOSR over other methods on CUB-200-2011. We set $P=0.99, \beta=1$ for these experiments. Despite the remarkable capability of CutMix, the combination of our method and CutMix still leads to better performance. The same improvement also happens to the combination of our method and other widely used techniques.

\begin{table}[h]
	\begin{center}
		\begin{small}
			\begin{tabular}{lllc}
				\toprule[1.3pt]
				\midrule
				Backbone                      & Method            &Acc. & + Ours \\ 
				\midrule
				\multirow{5}{*}{ResNet-50}  & +Cutout            & 85.85   & \textbf{86.22} \\    
				& +CutMix            & 86.19   & \textbf{86.73} \\ 
				\cmidrule(r){2-4}
				& +SOSR              &85.94    &--\\
				& +LabelSmoothing    & 85.90   & \textbf{86.22} \\ 
				& +ConfidencePenalty & 85.67   & \textbf{86.10} \\
				\midrule
				\bottomrule[1.3pt]
			\end{tabular}
		\end{small}
	\end{center}
	\caption{Accuracy ($\%$) over the combination of our method and other techniques on CUB-200-2011. ResNet-50 denotes the ResNet-50 network pre-trained with the regular process on ImageNet. We list the performance of SOSR itself for easier comparison.}
	\label{combinationoncub}
\end{table}

We draw attention to the fact that in spite of the similarity our method shares with Label Smoothing and Confidence Penalty, our method outperforms the other two methods when applied alone. More importantly, SOSR and this two similar methods are complementary, as evidenced by experimental results on a series of datasets.

\subsection{Visualizations of 2-D Features}
\begin{figure}[h]
	\begin{center}
		\includegraphics[width=0.97\columnwidth]{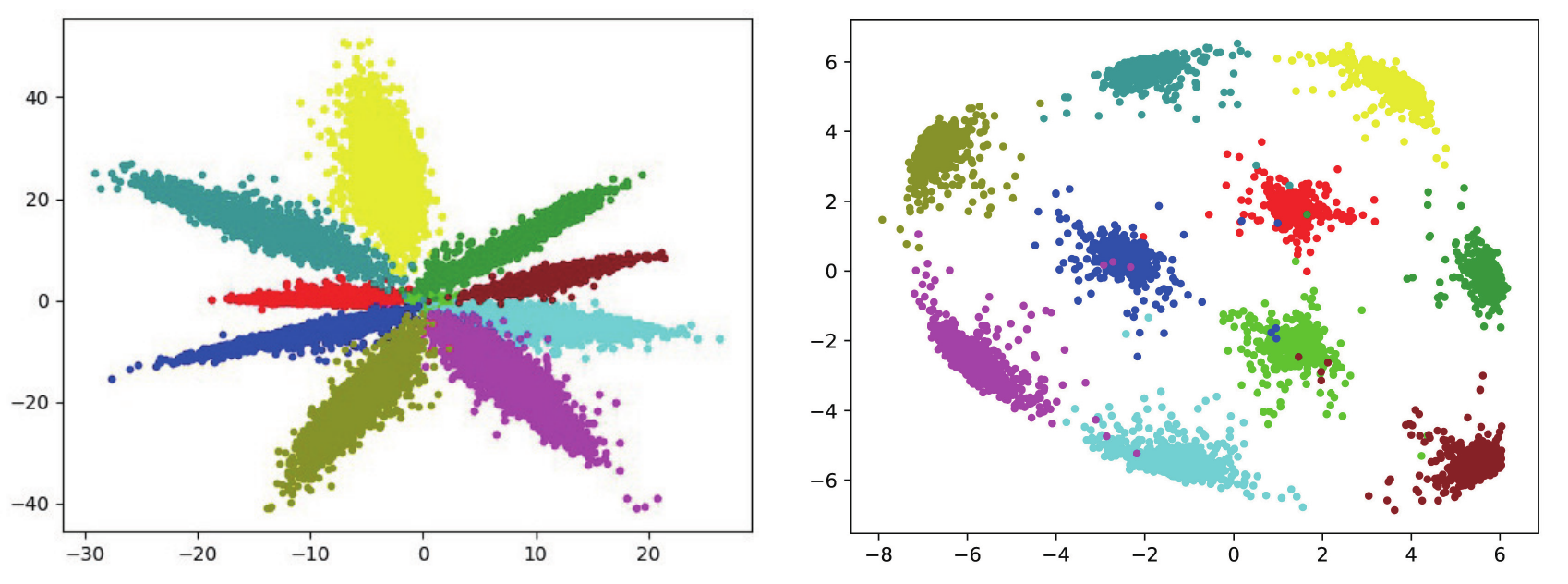} 
	\end{center}	
	\caption{Visualizations of 2-D feature distribution with ResNet-110. Left plot: baseline; Right plot: our method.}
	\label{featuredistribution}
\end{figure}

To better explain the effect of Selective Output Smoothing Regularization, we visualize how our method changes the representations learned by the model. As done by~\cite{wen2016discriminative}, we modify the ResNet-110 network on CIFAR-10 by reducing the output number of the last fully connected layer to 2 so that we can directly plot the features on 2-D coordinate axis for visualization. In Figure~\ref{featuredistribution}, we show the results of 2-D feature distribution of the models with and without our method.

From Figure~\ref{featuredistribution}, we can observe that the deeply learned features are separable successfully by the baseline model, demonstrating the effect of cross-entropy loss. Yet there is also a drawback that features are not discriminative enough, since a whole class of features are confused among the other features\footnote{The light green points in the center are completely flooded with points from the other 9 classes, as shown in baseline visualizations.}. In contrast to the baseline visualization, the model trained by our method shows a qualitatively different clustering of features. All the features from 10 classes are discernible and easy to be distinguished from each other, as evidenced by the clearer boundaries among features. We assume that it is because our method encourages each class to be distant from all the other classes. The apparent differences between the two visualizations demonstrate that SOSR helps obtain more highly discriminative features and thereby enlarge the inter-class feature differences.

\subsection{Object Detection in Pascal VOC}
In this section, we show Selective Output Smoothing Regularization can also be applied to training object detector in object detection task. Specifically, we want to examine whether models pre-trained with our method achieve better performance in Pascal VOC~\cite{everingham2010pascal} benchmark. We use RetinaNet\cite{lin2017focal} framework that runs convolutionally on Feature Pyramid Networks~\cite{lin2017feature} to localize and classify objects. We use fine-tuned ResNet-50 as our backbone architecture and follow the same anchor definition in~\cite{lin2017focal} to build FPNs. The models are trained with the union of Pascal VOC2007 and VOC2012 trainval data and evaluated on Pascal VOC2007 test data. Due to lack of resources, we curtail the number of training epochs to 30. We initialize the learning rate to 1e-5 and decrease it by 0.1 after the 3rd epoch if the loss still hasn't improved then. As seen in Table~\ref{objectdetection}, the SOSR pre-trained model significantly surpasses the vanilla ImageNet pretrained model in terms of mAP metric by 0.97$\%$, demonstrating that the superiority of our method in learning discriminative features can effectively lead to improvement in performance for object detection.

\begin{table}[h]
	\begin{center}
		\begin{small}
			\begin{tabular}{p{5.3cm}p{1.6cm}}
				\toprule[1.3pt]
				\midrule
				Backbone Network &  mAP\\
				\midrule
				pre-trained ResNet-50-FPN            & 70.17{\scriptsize $\pm$0.02}\\
				SOSR pre-trained ResNet-50-FPN        & 71.14{\scriptsize $\pm$0.06}\\
				\midrule
				\bottomrule[1.3pt]
			\end{tabular}
		\end{small}
	\end{center}
	\caption{Impact of Selective Output Smoothing Regularization on transfer learning of pretrained model to object detection.}
	\label{objectdetection}
\end{table}

%-------------------------------------------------------------------------
\section{Related Work}
\noindent\textbf{Regularization:} Regularization methods~\cite{huang2016deep,gastaldi2017shake,yamada2019shakedrop,ghiasi2018dropblock,zhang2021delving} have long been used in practice to mitigate overfitting. These works usually focus on adding randomness to keep the model from fitting too well in training data. Label Smoothing~\cite{szegedy2016rethinking} believes the model overfits because the model becomes too confident about its predictions and thereby modifies the label distribution to prevent the network from assigning full probability to each training example. Similarly, Confidence Penalty~\cite{pereyra2017regularizing} addresses this problem by constituting a regularization term that preventing peaked distribution. These two methods emphasize on avoiding the model from placing all probability on a single class in the training set, while our method manages to provide a mechanism to address the resultant negative effect produced by this situation.

\noindent\textbf{Data Augmentation:} It has long since been proved that data augmentation methods~\cite{zhong2020random,cubuk2018autoaugment,lim2019fast,xie20cut-thumbnail} can be used to improve the network’s robustness and overall performance It aims to enlarge the training set and introduce new knowledge by making minor alterations to samples. Cutout~\cite{devries2017improved} considers applying noise by masking out regions on inputs. CutMix~\cite{yun2019cutmix} removes some pixels and replaces them with a patch from another image. Due to the effective use of training pixels, CutMix achieves enormous improvements over the baselines and even some state-of-the-art performance. Despite the powerful ability of Cutout and CutMix, SOSR is still able to make further improvements when combining with them.

%-------------------------------------------------------------------------
\section{Conclusion}
In this paper, we introduce a simple but effective regularization method called Selective Output Smoothing Regularization. It acts like a “plug-and-play component” that can be incorporated into most CNN-based projects. Our method consistently outperforms the baseline in various experiment setups. Extensive experiments also show that our method can make further improvements even when combined with other techniques. Furthermore, simply using SOSR pre-trained model as the initialized backbone shows a nice boost in object detection.

\bibliographystyle{named}
\bibliography{ijcai22}

\end{document}